\documentclass[11pt]{article}

\usepackage[final]{acl}
\usepackage{times}
\usepackage{latexsym}
\usepackage[T1]{fontenc}
\usepackage[utf8]{inputenc}
\usepackage{microtype}
\usepackage{inconsolata}
\usepackage{graphicx}

\usepackage[utf8]{inputenc}
\usepackage[T1]{fontenc}
\usepackage{url}
\usepackage{booktabs}
\usepackage{amsfonts}
\usepackage{nicefrac}
\usepackage{microtype}
\usepackage{xcolor}
\usepackage{amsmath, amssymb, amsthm}
\usepackage{bm}
\usepackage{multirow}
\usepackage{graphicx}
\usepackage{caption}
\usepackage{subcaption}
\usepackage{enumitem}
\usepackage[most]{tcolorbox}
\usepackage{makecell}
\usepackage{tabularx}
\usepackage{array}
\usepackage{float}
\usepackage[flushleft]{threeparttable}
\usepackage[capitalize,noabbrev]{cleveref}
\usepackage[ruled,vlined,linesnumbered]{algorithm2e}
\usepackage{placeins}
\usepackage{listings}
\usepackage{inconsolata}
\usepackage{xspace}
\usepackage{newunicodechar}
\usepackage{pifont}
\usepackage[normalem]{ulem}
\usepackage{colortbl}
\newunicodechar{ }{\,}
\SetKwInput{Input}{Input}
\SetKwInput{Output}{Output}
\SetKw{Initialize}{Initialize}
\DontPrintSemicolon
\SetKwBlock{Block}{\{}{\}}

\newcommand{\std}[1]{{\scriptsize$\pm$#1}}

\definecolor{blue_cblind}{HTML}{1A85FF}

\definecolor{overallbg}{RGB}{218, 230, 252}  

\definecolor{orangesignal}{HTML}{FE6100}
\definecolor{highlightsignal}{HTML}{FFB000}
\definecolor{purplesignal}{HTML}{785ef0}
\definecolor{bluesignal}{HTML}{648fff}
\definecolor{pinksignal}{HTML}{FFB002}
\definecolor{citationcolor}{HTML}{648fff} 
\definecolor{linkcolorcustom}{HTML}{648fff}

\hypersetup{ 
    colorlinks=true,       
    linkcolor=pinksignal,
    citecolor=citationcolor, 
    urlcolor=linkcolorcustom 
}
\usepackage{titletoc}

\title{Pivot-SD: Efficient Self-Distillation for\\ Masked Diffusion Language Models}

\author{
  Seo Hyun Kim\textsuperscript{1,2,*\textdaggerdbl}%
  \thanks{Equal contribution.
  \textdagger Corresponding authors.
  \textdaggerdbl Work done as visiting researchers at the University of Toronto
  and the Vector Institute.} \quad
  Sunwoo Hong\textsuperscript{1,2,*\textdaggerdbl} \quad
  Younwoo Choi\textsuperscript{2} \quad
  Chen-Hao Chao\textsuperscript{2} \\[2pt]
  \bfseries
  Se-Young Yun\textsuperscript{1,\textdagger} \quad
  Rahul G. Krishnan\textsuperscript{2,\textdagger} \\[4pt]
  \textsuperscript{1}KAIST AI \quad
  \textsuperscript{2}University of Toronto \& Vector Institute \\
  \texttt{\{shkimsally, sunwoo\}@kaist.ac.kr} \\[3pt]
  \href{https://sunwoohong.github.io/pivot-sd/}{\texttt{sunwoohong.github.io/pivot-sd}}
}

\begin{document}
\maketitle

\begin{abstract}

Masked diffusion language models (dLMs) offer a promising parallel alternative to autoregressive models for complex reasoning. However, they face a distinct credit-assignment challenge, since a few commitments during denoising sharply reduce the uncertainty over the remaining masked positions and shape much of the response. Most post-training recipes for dLMs do not use this signal to decide which tokens to train on: they typically train on the final text or assign rewards to whole denoising steps, rather than selecting the individual commitments that shape the response. We introduce \textbf{Pivot-SD}, an efficient offline self-distillation framework that supervises only these high-impact commitments (pivots). Pivot-SD selects pivots using an information-gain metric measuring uncertainty reduction over the remaining masked positions. Pivots from successful trajectories are trained with cross-entropy, and pivots from failed trajectories with targeted unlikelihood, leaving the rest of the failed trajectory untouched. Using only 200 questions and four rollouts each, Pivot-SD improves LLaDA-8B-Instruct over full-sequence SFT and budget-matched diffusion RL baselines across math and code benchmarks.

\end{abstract}

\section{Introduction}
\label{sec:introduction}

Post-training with supervised fine-tuning (SFT) and reinforcement learning (RL) has driven much of the recent progress of large language models on reasoning \citep{ouyang2022training, shao2024deepseekmathpushinglimitsmathematical, deepseek-r1}. Not every token in a model's output is equally informative for training. A few choices constrain much of what follows, while most tokens are largely determined by the context around them. Which tokens a model should learn from is therefore a central question for post-training.

For autoregressive models, prior work selects tokens with token-level statistics \citep{wang20258020rulehighentropyminority, lin2025criticaltokensmattertokenlevel}, and training on the selected tokens improves reasoning. These statistics are computed on the final text, which loses nothing in an autoregressive model because the generation order is fixed by position. Masked diffusion language models (dLMs) are different. They start from a fully masked response and, under the confidence-based samplers used in practice, fill in positions in the order the model becomes confident about them \citep{sahoo2024simple, ye2025dream7bdiffusionlarge, nie2026large}. A dLM trajectory therefore records which token was written at which position and at which step. We call each of these events a commitment.

Most post-training recipes for dLMs do not use this record to decide which tokens to train on. Many are borrowed from autoregressive modeling and train on the final text \citep{zhao2026d, wang2026d2improvedtechniquestraining, rojas2026improving}. As a result, a token that fixes much of the response while most positions are still masked gets the same weight as a token filled in at the very end. In a failed trajectory, correct intermediate steps are penalized along with the error. Others operate on intermediate denoising steps \citep{chen2025reasoningdiffusionlargelanguage, zhan2026simplepolicygradientsreasoning, tang2026wd, xie-etal-2026-advancing, oba2026diffusionstatepolicyoptimizationmasked}, but they use them to guide online RL updates.

\begin{figure}[t]
    \centering
    \includegraphics[width=\columnwidth]{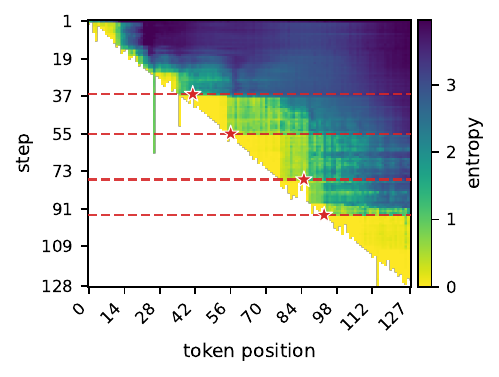}
    \caption{\textbf{Predictive Entropy and Pivot Selection.} An entropy heatmap of a masked denoising trajectory for a MATH problem (showing the first 128 tokens). The x-axis represents the token position, and the y-axis denotes the denoising step progressing downward. The color scale indicates the predictive entropy of the active masked positions. Red dashed lines and stars denote the critical denoising steps and the specific pivot tokens selected by our Information Gain metric, respectively.}
    \label{fig:entropy_heatmap}
\end{figure}

In these trajectories, the entropy over the masked positions drops markedly at a few denoising steps and changes little at the others (Figure~\ref{fig:entropy_heatmap}). We call the tokens committed at these steps pivots. We score each step by its information gain, the normalized drop in entropy over the masked positions caused by the step's commitment. Related signals have been used at inference time to choose which positions to unmask \citep{NEURIPS2025_8514c1f1, yang2026improvingsamplingmaskeddiffusion}. We use information gain to choose which tokens to train on. In a correct trajectory, pivots most shaped its path and are the tokens to learn from. In an incorrect trajectory, they most constrained the failed path, which makes them the natural targets for a negative update.

We introduce Pivot-SD, a self-distillation method that trains a dLM only on its pivots. Pivot-SD samples trajectories once from the base model, extracts their pivots, and stores the partially masked state in which each pivot was committed. Training replays these states, so the model predicts each pivot token in the exact state where it was originally committed. Whether the final answer is correct sets the direction of each update. Pivots from correct trajectories are trained with cross-entropy to raise their probability, and pivots from incorrect ones with unlikelihood to lower it \citep{welleck2019neuraltextgenerationunlikelihood, ni2025offlinelearningforgettingreasoning}.

Pivot-SD computes its loss on only 10 tokens per trajectory, under 4\% of the 256-token response budget. With 200 training questions, it gives LLaDA-8B-Instruct \citep{nie2026large} the highest mean accuracy on four math and code benchmarks. It outperforms full-sequence SFT and diffusion RL, including RL runs trained on ten times as many questions for five times as many steps, while using one-sixth to one-eighth of their wall-clock time. It also improves a second backbone, Dream-7B \citep{ye2025dream7bdiffusionlarge}, and has the best average out-of-domain accuracy. Ablations confirm that the gains depend on which tokens are selected, since supervising random steps or every masked token lowers accuracy.

Our contributions are as follows:
\begin{itemize}
    \item We identify \textbf{pivots} with a signal specific to dLMs that autoregressive generation does not provide, namely how much each denoising step reduces the uncertainty over the masked positions. Pivots are the tokens committed at the steps where this reduction is largest.
    \item We introduce \textbf{Pivot-SD}, a self-distillation method that replays the partially masked states in which the pivots were committed. It applies cross-entropy to the pivots of successful trajectories and unlikelihood to those of failed ones.
    \item We show that supervising under 4\% of the response budget is enough. With only 200 training questions, Pivot-SD achieves the highest mean accuracy among the compared methods on math and code benchmarks. It outperforms online RL baselines, including runs with ten times as many questions and five times as many training steps, while using about 40\% fewer FLOPs than budget-matched RL.
\end{itemize}
\section{Related Work}
\label{sec:related_work}

\paragraph{Diffusion language models.}
Diffusion models have been studied as alternatives to autoregressive language modeling. Early work explored continuous diffusion over word embeddings \citep{li2022diffusionlm}, and D3PMs extended denoising diffusion to discrete state spaces \citep{austin2021structured}. Later work improved discrete and masked diffusion with score-entropy objectives and masked training recipes \citep{3692070.3693403, sahoo2024simple}, and LLaDA scaled the paradigm to an 8B instruction-following model trained with forward masking and reverse masked-token prediction \citep{nie2026large}. In these models generation passes through a sequence of partially masked states instead of a fixed left-to-right order, and Pivot-SD takes its supervision from this sequence of states.

\paragraph{Reasoning with masked diffusion language models.}
Several recent methods adapt these models for reasoning. d1 combines masked SFT with diffu-GRPO, an RL algorithm for masked diffusion \citep{zhao2026d}. DCoLT treats intermediate reverse-diffusion steps as latent thinking actions and optimizes trajectories with outcome-based RL \citep{huang2026reinforcing}. d2 proposes policy-gradient techniques that improve reasoning without supervised fine-tuning \citep{wang2026d2improvedtechniquestraining}, and GDPO revisits likelihood estimation and uses variance-reduced ELBO-based policy optimization \citep{rojas2026improving}. AGRPO treats each unmasking step as an action and estimates the policy gradient from
sampled steps \citep{zhan2026simplepolicygradientsreasoning}.
These methods rely on online RL. Pivot-SD instead samples
trajectories once from a frozen model and converts trajectory outcomes into localized supervision over selected denoising commitments. This gives a stable offline alternative that can use both successful and failed trajectories.

\paragraph{Process supervision and reasoning credit assignment.}
A common approach to improving mathematical reasoning is to supervise intermediate steps. \citet{lightman2024lets} show that step-level feedback can outperform outcome supervision on MATH, and Math-Shepherd automatically constructs process labels and trains process reward models \citep{wang-etal-2024-math}. These methods score textual reasoning steps in autoregressive solutions.

Pivot-SD shares the goal of fine-grained credit assignment but changes the unit of supervision. Instead of scoring a sentence or derivation step, it supervises a denoising commitment $(t, p, y_p)$ inside a masked diffusion trajectory. This matters because masked diffusion models commit tokens out of order, and a single commitment can reshape the distribution over many remaining positions.

\paragraph{Token importance and weighted fine-tuning.}
A related direction selects or reweights tokens during training. For autoregressive models, \citet{wang20258020rulehighentropyminority} restrict policy-gradient updates to high-entropy tokens, and \citet{lin2025criticaltokensmattertokenlevel} penalize the critical tokens of incorrect trajectories. For diffusion language models, GIFT assigns
entropy-based importance weights to tokens \citep{xu2026giftguidedimportanceawarefinetuning}, showing that emphasizing some tokens helps fine-tuning. These methods score positions of the final sequence. Pivot-SD scores a commitment at the step where it was made, by its effect on the positions that are still masked, which the final sequence does not record. Our ablations compare pivot-only supervision, selected-step uniform supervision, entropy-selected pivots, and random pivots.

\paragraph{Positioning of our work.}
Table~\ref{tab:credit-landscape-full} in Appendix~\ref{app:credit-landscape} compares these methods by the unit they supervise and the signal that localizes it. Process supervision scores textual steps \citep{lightman2024lets, wang-etal-2024-math}, GIFT reweights final-sequence positions \citep{xu2026giftguidedimportanceawarefinetuning}, step-aware diffusion RL assigns rewards over denoising intervals \citep{chen2025reasoningdiffusionlargelanguage, xie-etal-2026-advancing, oba2026diffusionstatepolicyoptimizationmasked}, and answer-aware selection targets the terminal answer block \citep{horvitz2026rethinkingreasoningmdlmsearly}. Pivot-SD trains only on selected commitments inside the trajectory and takes both positive and negative targets from verifier outcomes, without a process reward model or step-level labels.
\newcommand{\ours}{Pivot-SD}
\newcommand{\Resp}{\mathrm{Resp}}
\newcommand{\Extract}{\mathrm{Extract}}
\newcommand{\Ans}{\mathrm{Ans}}
\newcommand{\ind}{\mathbb{I}}

\begin{figure*}[htbp]
    \centering
    \includegraphics[width=0.95\linewidth]{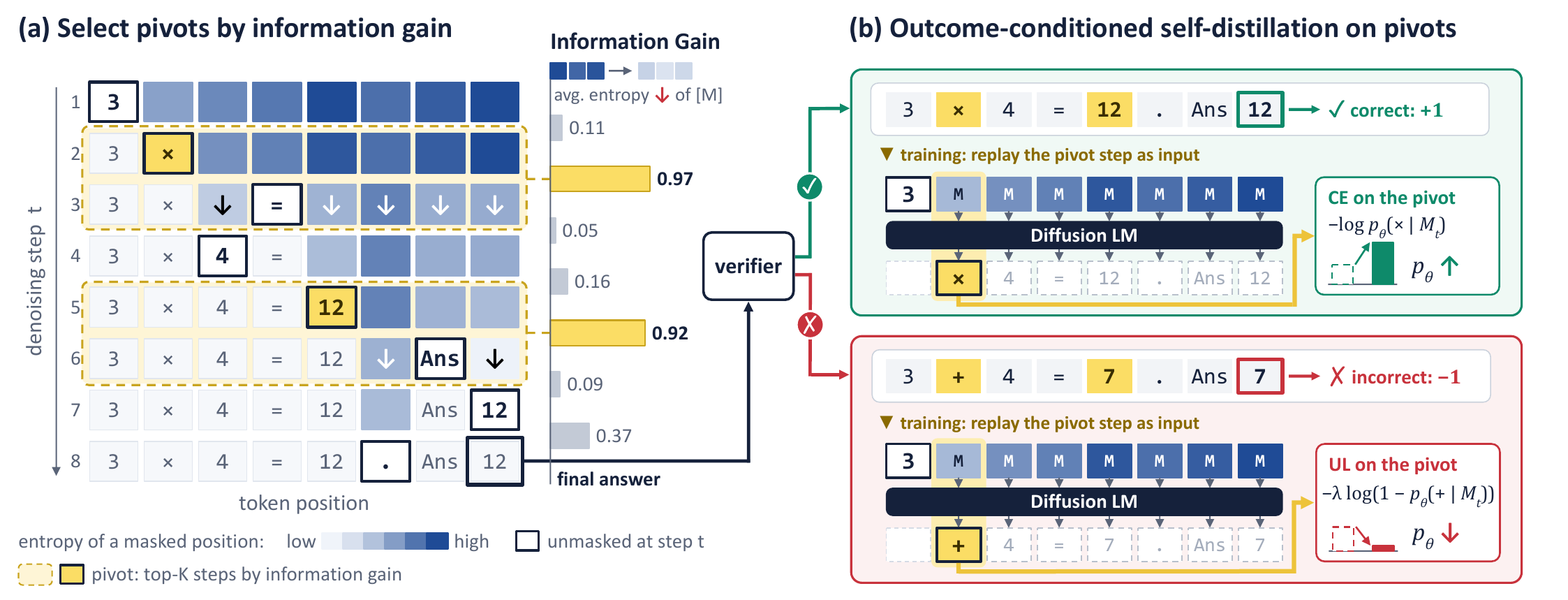}
    \caption{\textbf{Illustrative Overview of Pivot-SD.} (a) In a trajectory sampled from $\theta_0$, steps whose commitment most reduces the average entropy of the remaining masked positions are selected as pivots (yellow). (b) The final answer determines the sign: for each pivot, its masked state $M_t$ is replayed and only the pivot token is trained, with cross-entropy if the answer is correct and unlikelihood if it is wrong. All other positions receive no loss.}
    \label{fig:pivot_sd_main}
\end{figure*}

\section{Methodology}
This section introduces notation and the SFT baseline (\S\ref{sec:preliminaries}), the pivot selection rule and training objective of Pivot-SD (\S\ref{sec:pivot-sd-framework}), the full algorithm (\S\ref{sec:algo}), and its relation to SFT and online RL (\S\ref{sec:relation_to_sft_rl}).
\subsection{Preliminaries and Problem Setup}
\label{sec:preliminaries}
\paragraph{Setup}
Let $x$ represent a conditioning prompt, and let $y = (y_1, \ldots, y_L)$ denote a target response sequence of length $L$. A masked diffusion language model generates $y$ conditioned on $x$ by iteratively denoising an initially fully masked response sequence over $T$ steps. Let $M_t$ represent the complete sequence state at denoising step $t$ (consisting of the unmasked prompt $x$ concatenated with the partially masked response), indexed in the sampler's reverse chronological order.
To formalize the active response positions, let $\mathcal{I}_{\text{resp}}$ denote the set of coordinate indices corresponding exclusively to the response tokens. The active response positions that remain masked at step $t$ are defined as
\begin{equation}
\mathcal{B}_t = \{i \in \mathcal{I}_{\text{resp}} : M_{t,i} = [\mathrm{MASK}]\}.
\label{eq:active-masked-set}
\end{equation}
At each denoising step, the model parameterizes a distribution over the clean token for each active masked position:
\begin{equation}
P_\theta(x_0^{(i)} \mid M_t), \qquad i \in \mathcal{B}_t,
\label{eq:model-prediction}
\end{equation}
where $x_0^{(i)}$ is the clean-token random variable at response position $i$, and its final sampled value in the finalized trajectory is $y_i$.

\paragraph{Standard SFT}
Standard masked SFT trains the parameters $\theta$ by minimizing the uniform per-position cross-entropy loss:
\begin{equation}
\mathcal{L}_{\mathrm{SFT}}(\theta) = -\mathbb{E}_{((x,y), t, M_t)} \left[ \sum_{i \in \mathcal{B}_t} \log P_{\theta}(y_i \mid M_t) \right].
\label{eq:sft}
\end{equation}
Eq.~\eqref{eq:sft} weights every masked position equally, whether it is a trivial fill or a commitment that determines the rest of the response. Pivot-SD instead supervises a small set of selected commitments and sets the sign of each update by whether the trajectory succeeded.

\subsection{The Pivot-SD Framework}
\label{sec:pivot-sd-framework}
Pivot-SD has three parts: extracting candidate commitments from sampled trajectories (\S\ref{sec:trajectory-generation}), selecting pivots by information gain (\S\ref{sec:information-gain-pivot-selection}), and the training objective (\S\ref{sec:training-objective}). Figure~\ref{fig:pivot_sd_main} gives an illustrative overview.

\subsubsection{Trajectory Generation and Candidate Pivots}
\label{sec:trajectory-generation}
Sampling a complete response from the frozen base model $\theta_0$ yields a discrete denoising trajectory:
\begin{equation}
\tau = \{(M_t, \mathcal{B}_t, \mathcal{U}_t)\}_{t=1}^{T},
\label{eq:traj}
\end{equation}
where $\mathcal{U}_t \subseteq \mathcal{B}_t$ is the specific subset of positions selected for unmasking and commitment by the sampler at step $t$. A position $p$ is committed at step $t$ when the sampler selects $p \in \mathcal{U}_t$ and writes the sampled token value $y_p$ into the sequence state.
For every committed position $p \in \mathcal{U}_t$, the trajectory generates a candidate pivot $c$, which we group into a trajectory-wide candidate set:
\begin{equation}
\mathcal{C}(\tau) = \{(t, p, y_p) : p \in \mathcal{U}_t\}.
\label{eq:candidate-set}
\end{equation}
The triple in Eq.~\eqref{eq:candidate-set} records which token was committed, at which position, and at which step, so the supervised object is the event of committing $y_p$ at $p$ under state $M_t$. The same token at the same position can matter little late in denoising and a great deal earlier, when many neighboring positions are still masked. Token-level SFT supervises only $(p,y_p)$, and trajectory-level RL supervises only the outcome.

\subsubsection{Information-Gain Pivot Selection}
\label{sec:information-gain-pivot-selection}
A pivot is a commitment after which the frozen model becomes much more certain about the positions that are still masked. We therefore score each denoising step by how much the entropy of those positions drops across the step. For a masked state $M$ and position $i$, the per-position entropy is
\begin{equation}
h_{\theta_0}(i \mid M) = H\left(P_{\theta_0}(x_0^{(i)} \mid M)\right).
\label{eq:entropy-def}
\end{equation}
Let $\mathcal{A}_t \subseteq \mathcal{B}_t$ be the positions the sampler can unmask at step $t$, and let $M_t^{[\mathcal{U}_t]}$ denote $M_t$ with the tokens committed at step $t$ written in. $M_t^{[\mathcal{U}_t]}$ is the state the sampler evaluates at the next step. Over the positions that remain masked after step $t$, we compare the entropy before and after the commitment:
\begin{equation}
H_{\mathrm{pre}}(t) = \sum_{i \in \mathcal{A}_t \setminus \mathcal{U}_t} h_{\theta_0}(i \mid M_t),
\label{eq:h-pre}
\end{equation}
\begin{equation}
H_{\mathrm{post}}(t) = \sum_{i \in \mathcal{A}_t \setminus \mathcal{U}_t} h_{\theta_0}\left(i \mid M_t^{[\mathcal{U}_t]}\right).
\label{eq:h-post}
\end{equation}
Both sums exclude the committed positions, so their difference measures how much the commitment constrains the rest of the block. The number of positions in the sum shrinks as decoding proceeds, so a raw difference would favor early steps simply because more positions are still masked. We therefore divide by that count:
\begin{equation}
g(t) = \frac{H_{\mathrm{pre}}(t) - H_{\mathrm{post}}(t)}{\lvert \mathcal{A}_t \setminus \mathcal{U}_t \rvert},
\label{eq:ig-score}
\end{equation}
so that $g(t)$ is the average entropy reduction per remaining masked position, which is comparable across steps. Steps at which $\mathcal{A}_t$ changes are excluded, since $H_{\mathrm{pre}}$ and $H_{\mathrm{post}}$ would then be measured over different sets of positions.

For each trajectory we keep the $K$ highest-scoring steps and supervise every token committed at those steps; the sampler commits one token per step in our setting (Appendix~\ref{sec:app-experimental-details}), so this yields $K$ pivots per trajectory. We set $K=10$ before running experiments, and results change little for $5 \le K \le 20$ (Appendix~\ref{sec:app-ksweep}). The score only selects pivots and does not weight the loss.

Both entropy terms come from forward passes that the sampler already runs at step $t$ and at the following step, so scoring adds no model calls to generation.

Figure~\ref{fig:entropy_heatmap} shows a real trajectory of the base model on a MATH problem, where the selected steps (red dashed lines) coincide with sharp drops in the entropy of the remaining positions.

\begin{table*}[t]
\centering
\small 
\caption{
\textbf{Accuracy of LLaDA-8B-Instruct on math and code benchmarks} at generation lengths of 256 and 512. Budget-matched RL variants use the same 200 questions and 1,000 optimizer steps as Pivot-SD. Methods marked with $\ddag$ are averaged over three fully re-randomized runs at 256 tokens. Standard deviations are in Appendix~\ref{app:multiseed}.
}
\label{tab:llada-main}
\begin{tabular}{lcccccccc}
\toprule
\textbf{Method}
& \multicolumn{2}{c}{\textbf{MATH}}
& \multicolumn{2}{c}{\textbf{GSM8K}}
& \multicolumn{2}{c}{\textbf{HumanEval+}}
& \multicolumn{2}{c}{\textbf{MBPP+}} \\
\cmidrule(lr){2-3} \cmidrule(lr){4-5} \cmidrule(lr){6-7} \cmidrule(lr){8-9}
& 256 & 512 & 256 & 512 & 256 & 512 & 256 & 512 \\
\midrule
Base 
& 31.40 & 38.8 & 75.13 & 83.40 & 32.32 & \textbf{42.07} & 43.12 & 44.44 \\
SFT-GT$^{\ddag}$ 
& 35.47 & 38.6 & 76.94 & 83.17 & 33.94 & 36.59 & 46.32 & 42.86 \\
SFT-SD$^{\ddag}$ 
& 31.73 & 38.6 & 73.90 & 81.96 & 35.78 & 40.85 & 45.92 & 44.97 \\
diffu-GRPO (budget-matched)$^{\ddag}$
& 33.47 & 38.2 & 76.04 & 82.11 & 32.52 & \textbf{42.07} & 43.39 & 45.24 \\
diffu-GRPO 
& 35.20 & 40.6 & 77.63 & 83.62 & 34.15 & 40.85 & 42.59 & 43.65 \\
wd1++ (budget-matched)$^{\ddag}$
& 33.27 & 39.2 & 76.93 & 82.71 & 33.54 & 40.24 & 43.12 & 43.92 \\
wd1++ 
& 35.40 & 39.4 & 76.65 & 82.71 & 34.15 & 38.41 & 44.97 & 44.44 \\
\midrule
\rowcolor{overallbg}
\textbf{Pivot-SD (ours)}$^{\ddag}$
& \textbf{37.47} & \textbf{42.6} & \textbf{79.51} & \textbf{84.38} & \textbf{40.43} & 41.46 & \textbf{47.12} & \textbf{48.15} \\
\bottomrule
\end{tabular}
\end{table*}

\subsubsection{Training objective}
\label{sec:training-objective}
Each selected pivot is mapped to a trajectory outcome score $s(\tau) \in \{+1, -1\}$, which evaluates the correctness of the final response sequence $y$ generated by trajectory $\tau$:
\begin{equation}
    s(\tau)
    =
    \begin{cases}
    +1, & \text{if } y \text{ is correct}, \\
    -1, & \text{otherwise}.
    \end{cases}
\label{eq:trajectory-outcome}
\end{equation}
Every chosen pivot yields a training tuple $z = (M_t, y, p, y_p, s(\tau))$. For a selected pivot token $y_p$, we define the positive cross-entropy loss as
\begin{equation}
\ell^{+}_p(\theta; M_t, y_p) = -\log P_\theta(y_p \mid M_t),
\label{eq:pivot-ce}
\end{equation}
and the token-level unlikelihood penalty as
\begin{equation}
\ell^{-}_p(\theta; M_t, y_p) = -\log \left(1 - P_\theta(y_p \mid M_t)\right).
\label{eq:pivot-ul}
\end{equation}
During training, the predicted probability $P_\theta(y_p \mid M_t)$ in Eq.~\eqref{eq:pivot-ul} is clipped away from $1$ to maintain numerical stability.
Failed trajectories often contain valid syntax or correct intermediate algebra. Applying unlikelihood only at the selected pivots leaves these parts of a failed trajectory untouched (Figure~\ref{fig:pivot_sd_main}(b)). The loss for a single tuple $z$ is

\begin{equation}
\begin{aligned}
    L_{\mathrm{pivot}}(z)
    &=
    \mathbf{1}[s(\tau)=+1]\,
    \ell^{+}_p(\theta; M_t, y_p)
    \\
    &\quad+
    \lambda_{\mathrm{neg}}\,
    \mathbf{1}[s(\tau)=-1]\,
    \ell^{-}_p(\theta; M_t, y_p),
\end{aligned}
\label{eq:pivot-loss}
\end{equation}

where $\lambda_{\mathrm{neg}}$ is a hyperparameter scaling the unlikelihood penalty.
Collecting the selected pivots over all training prompts $\mathcal{X}$ and their trajectories gives the offline dataset $\mathcal{D}_{\mathrm{pivot}}$, and we minimize
\begin{equation}
\min_{\theta} \frac{1}{|\mathcal{D}_{\mathrm{pivot}}|} \sum_{z \in \mathcal{D}_{\mathrm{pivot}}} L_{\mathrm{pivot}}(z).
\label{eq:pivot-objective}
\end{equation}
We initialize $\theta$ from the base model $\theta_0$ and optimize Eq.~\eqref{eq:pivot-objective} using standard gradient-based optimization.

\begin{table*}[t]
\centering
\small
\renewcommand{\arraystretch}{1.08}
\caption{
\textbf{Ablation study of pivot selection and credit assignment.}
``IG'' denotes denoising states selected by our information-gain criterion.
``All'' applies the loss to all supervised masked tokens in the selected state, while
``Pivot'' applies the loss only to the selected pivot token.
CE denotes cross-entropy and UL denotes unlikelihood.
Methods marked with $\ddag$ are averaged over three fully re-randomized runs.
Standard deviations are in Appendix~\ref{app:multiseed}.
}
\label{tab:ablation-compact}
\begin{tabular}{l l l l l c c c c}
\toprule
\textbf{Method}
& \textbf{State}
& \textbf{Target}
& \textbf{Pos.}
& \textbf{Neg.}
& \textbf{MATH}
& \textbf{GSM8K}
& \textbf{HumanEval+}
& \textbf{MBPP+} \\
\midrule
Positive All-token
& IG
& All
& CE
& -
& 31.60
& 74.37
& 36.59
& 39.68 \\
Positive Pivot
& IG
& Pivot
& CE
& -
& 32.20
& 72.48
& 35.37
& 43.65 \\
Entropy pivot$^{\ddag}$
& Entropy
& Pivot
& CE
& -
& 32.27
& 74.75
& 35.57
& 44.80 \\
Pos+Neg All-token$^{\ddag}$
& IG
& All
& CE
& State-UL
& 35.00
& 74.44
& 36.80
& 44.54 \\
Random-Step Pivot$^{\ddag}$
& Random
& Pivot
& CE
& Pivot-UL
& 29.93
& 76.51
& 39.00
& 44.80 \\
Random-Token Pivot$^{\ddag}$
& IG
& Random
& CE
& Pivot-UL
& 34.87
& 76.47
& 34.78
& 45.24 \\
\midrule
\rowcolor{overallbg}
\textbf{\ours{} (ours)}$^{\ddag}$
& IG
& Pivot
& CE
& Pivot-UL
& \textbf{37.47}
& \textbf{79.51}
& \textbf{40.43}
& \textbf{47.12} \\
\bottomrule
\end{tabular}
\end{table*}

\subsection{Algorithmic Implementation}
\label{sec:algo}

\begin{algorithm}[t]
\small
\caption{\ours{}: Offline Outcome-Conditioned Pivot Self-Distillation}
\label{alg:pivot-sd}
\Input{Training prompts $\mathcal{X}$, frozen base model $\theta_0$, trajectories per prompt $G$, pivots per trajectory $K$}
\Output{Fine-tuned model $\theta$}
\Initialize pivot dataset $\mathcal{D}_{\mathrm{pivot}} \leftarrow \emptyset$\;
\ForEach{prompt $x \in \mathcal{X}$}{
  \For{$j \leftarrow 1$ \KwTo $G$}{
    Sample a denoising trajectory $\tau=\{(M_t,\mathcal{B}_t,\mathcal{U}_t)\}_{t=1}^{T}$ and response $y$ from $\theta_0$\;
    Assign trajectory outcome $s(\tau)$ using Eq.~\eqref{eq:trajectory-outcome}\;
    Construct candidate pivots $\mathcal{C}(\tau)=\{(t,p,y_p): p \in \mathcal{U}_t\}$\;
    \ForEach{denoising step $t$ of $\tau$}{
      Compute $g(t)$ using Eq.~\eqref{eq:ig-score}\;
    }
    Select the top-$K$ steps by $g(t)$\;
    \ForEach{$(t,p,y_p) \in \mathcal{C}(\tau)$ with $t$ among the selected steps}{
      Add $(M_t,y,p,y_p,s(\tau))$ to $\mathcal{D}_{\mathrm{pivot}}$\;
    }
  }
}
Fine-tune $\theta$ on $\mathcal{D}_{\mathrm{pivot}}$ using Eq.~\eqref{eq:pivot-objective}\;
\Return{$\theta$}
\end{algorithm}

Algorithm~\ref{alg:pivot-sd} summarizes the pipeline from trajectory sampling to fine-tuning.

\subsection{Relation to SFT and Online RL}
\label{sec:relation_to_sft_rl}

Like SFT, Pivot-SD trains offline with supervised losses. Like RL, it uses outcome labels from both successful and failed trajectories. It differs from both in where the update is applied. Each trajectory outcome is converted into $K$ state-conditioned targets: information gain chooses the commitments, the verifier chooses the direction of the update, and cross-entropy or unlikelihood is applied only at those commitments. This is what lets Pivot-SD learn from failed trajectories without penalizing their valid parts.

Because $\mathcal{D}_{\mathrm{pivot}}$ is sampled once from $\theta_0$ and stays fixed during training, hyperparameters can be tuned offline on the same dataset, and training cannot stall on zero-reward batches, which we observed for online RL at this budget (Appendix~\ref{sec:appendix_bootstrap}).
\newcommand{\tbd}{\textsc{TBD}}
\newcommand{\llada}{LLaDA-8B-Instruct}
\newcommand{\dream}{Dream-v0-Instruct-7B}
\newcommand{\dash}{--}

\newcolumntype{L}{>{\raggedright\arraybackslash}X}

\section{Experiments}
\label{sec:experiments}

We test whether pivot self-distillation improves masked diffusion language models under a small post-training budget, using \llada{} as the main backbone.

\paragraph{Training data.}
For each math reasoning dataset, we sample 200 training questions from the corresponding training split: GSM8K for GSM8K evaluation \citep{cobbe2021trainingverifierssolvemath} and MATH for MATH500 \citep{hendrycks2021measuring, lightman2024lets} evaluation. For code tasks, we sample 200 training questions from AceCode \citep{zeng2025acecoderacingcoderrl} and evaluate on HumanEval+ and MBPP+ \citep{chen2021evaluatinglargelanguagemodels, liu2023is,austin2021programsynthesislargelanguage}. For each training question, we generate four denoising trajectories from the frozen base model, for 800 trajectories per dataset. Each trajectory is labeled as successful or failed using the task verifier: exact-answer matching for math and unit-test execution for code.

\paragraph{Baselines.}
We compare against two offline SFT baselines and two online RL baselines. SFT-GT trains on ground-truth solutions with standard random masking. SFT-SD trains on the successful solutions from the same rollout pool as \ours{}, with ordinary masked SFT. The RL baselines are diffu-GRPO \citep{zhao2026d} and wd1++ \citep{tang2026wd}. wd1++ is the step-wise variant of wd1 proposed by \citet{tang2026wd}, which also trains on the intermediate clean completions produced at each denoising step. We use it because it is the stronger of their two variants. For both RL methods, the budget-matched variant uses the same 200 questions, four rollouts per question, and 1,000 optimizer steps as \ours{}, and the extended variant uses 2,000 questions and 5,000 steps.

\paragraph{Evaluation.}
All rollout-based methods collect training trajectories with a maximum generation length of 256 tokens, and SFT-GT trains on ground-truth solutions that fit the same budget. We evaluate on MATH500, GSM8K, HumanEval+, and MBPP+ with a shared decoding configuration and answer-extraction or unit-test pipeline. The 256-token setting matches training and is our main setting. The 512-token setting measures length transfer. Methods trained under the controlled budget are run three times with full re-randomization at 256 tokens. The main tables report means, and Appendix~\ref{app:multiseed} gives standard deviations and the full protocol.
Detailed experimental configurations and hyperparameters are provided in Appendix~\ref{sec:app-experimental-details}.

\subsection{Main Results}
\label{subsec:main-results}

Table~\ref{tab:llada-main} reports the main comparison. At the 256-token budget used to collect pivot supervision, \ours{} has the best mean on all four benchmarks, including against the 5,000-step RL runs. Over the strongest baseline on each benchmark, it gains 2.0 points on MATH (over SFT-GT), 1.9 on GSM8K (over 5,000-step diffu-GRPO), and 4.7 on HumanEval+ (over SFT-SD). On MBPP+ the margin over SFT-GT is 0.8 points, within one standard deviation (Appendix~\ref{app:multiseed}). SFT-SD trains on the same rollout pool with ordinary masked SFT, so its gap to \ours{} reflects pivot selection and negative supervision on identical data. Across training budgets from 50 to 2,000 questions, \ours{} stays best on HumanEval+ and is best on all four benchmarks at 2,000 questions (Appendix~\ref{sec:app-budget}).

At 512 tokens, \ours{} is best on MATH, GSM8K, and MBPP+. On HumanEval+ it scores 41.46, below the base model and budget-matched diffu-GRPO (both 42.07).

\begin{table}[t]
\centering
\small
\setlength{\tabcolsep}{4pt}
\renewcommand{\arraystretch}{1.1}
\caption{
\textbf{Compute accounting.} Wall-clock hours on two GPUs. \ours{} generates its trajectories once before training. The RL baselines generate rollouts during training. RL times are averaged across training domains.
}
\label{tab:compute}

\resizebox{\columnwidth}{!}{%
\begin{tabular}{l c c c c}
\toprule
\textbf{Method}
& \textbf{Steps}
& \textbf{Gen. (h)}
& \textbf{Train (h)}
& \textbf{Total (h)} \\
\midrule

diffu-GRPO (matched)
& 1000
& 0
& 5.4
& 5.4 \\

diffu-GRPO
& 5000
& 0
& 22.9
& 22.9 \\

wd1++ (matched)
& 1000
& 0
& 3.6
& 3.6 \\

wd1++
& 5000
& 0
& 17.3
& 17.3 \\

\midrule
\rowcolor{overallbg}
\textbf{\ours{} (ours)}
& 1000
& 2.5
& 0.3
& \textbf{2.8} \\

\bottomrule
\end{tabular}%
}
\end{table}

\begin{figure}[t]
    \centering
    \includegraphics[width=\columnwidth]{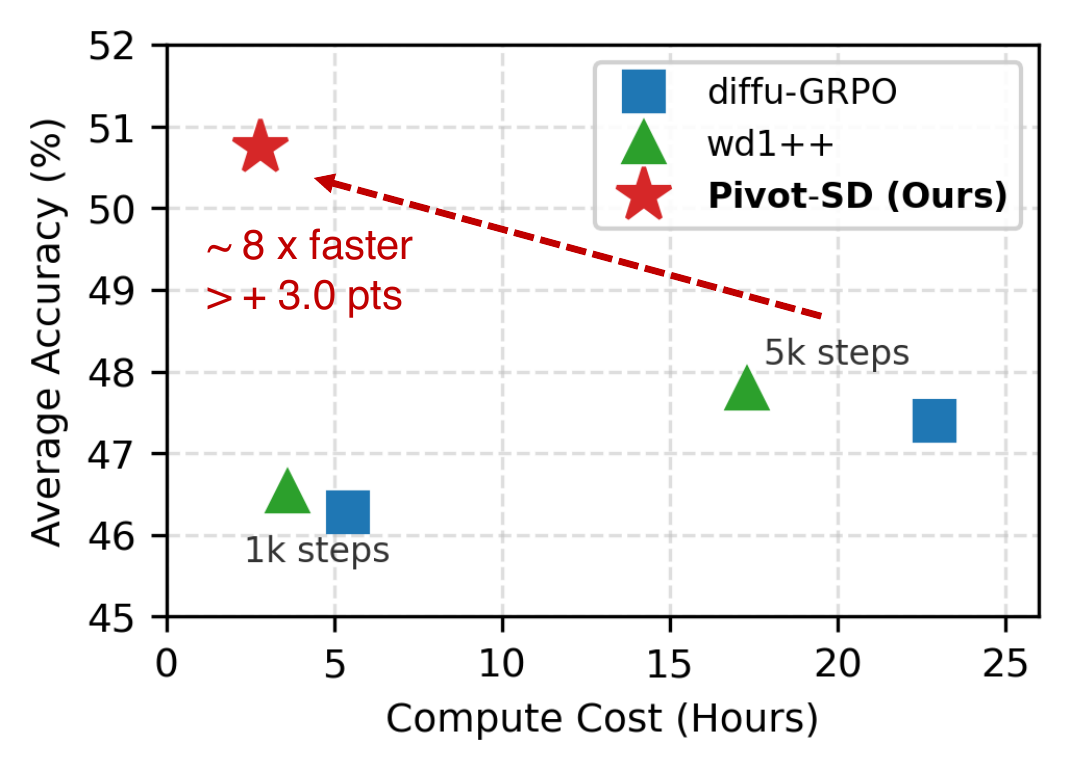}
    \caption{\textbf{Accuracy vs. compute cost.} Average accuracy versus total wall-clock hours on two GPUs. \ours{} is more than 3 points above the 5,000-step RL runs while using about one-eighth of the training time of 5,000-step diffu-GRPO.}
    \label{fig:acc_compute}
\end{figure}

\begin{table}[t]
\centering
\footnotesize
\setlength{\tabcolsep}{3pt}
\renewcommand{\arraystretch}{1.08}
\caption{
\textbf{Dream generalization results.} We keep the same baseline categories as in the LLaDA
experiments and report the same evaluation metrics.
}
\label{tab:dream-compact}
\begin{tabularx}{\columnwidth}{@{}Lcccc@{}}
\toprule
\textbf{Method}
& \textbf{MATH}
& \textbf{GSM}
& \textbf{HE}
& \textbf{MBPP} \\
\midrule
Base (Dream-Instruct-7B)
& 37.97 & 79.55 & 54.27 & 60.58 \\

SFT-GT
& 35.40 & 61.49 & 54.27 & 59.26 \\

SFT-SD
& 40.80 & 70.74 & 48.78 & 61.38 \\

\midrule
\rowcolor{overallbg}
\textbf{\ours{} (ours)}
& \textbf{42.20} & \textbf{81.88} & \textbf{58.54} & \textbf{61.64} \\
\bottomrule
\end{tabularx}
\end{table}

\begin{table}[t]
\centering
\footnotesize
\renewcommand{\arraystretch}{1.08}
\setlength{\tabcolsep}{4pt}
\caption{
\textbf{Average out-of-domain (OOD) performance.} Each column reports the mean accuracy over the benchmarks not used for training: MATH- and GSM8K-trained models are evaluated on the other three benchmarks, AceCode-trained models on MATH and GSM8K. Column composition therefore differs, and ``Overall'' is the micro-average over all source-target pairs rather than the mean of the three columns.
}
\label{tab:ood-avg}
\resizebox{\columnwidth}{!}{%
\begin{tabular}{l c c c c}
\toprule
\multirow{2}{*}{\textbf{Method}} 
& \multicolumn{3}{c}{\textbf{Training domain}} 
& \multirow{2}{*}{\textbf{Overall}} \\
\cmidrule(lr){2-4}
& \textbf{MATH} & \textbf{GSM8K} & \textbf{AceCode} & \\
\midrule
Base                 & 50.19          & 35.61          & 53.27          & 45.49          \\
SFT-GT               & \textbf{53.57} & 39.46          & 51.80          & 47.83          \\
SFT-SD               & 51.13          & 35.44          & 51.21          & 45.27          \\
diffu-GRPO (matched) & 50.89          & 36.45          & \textbf{55.49} & 46.62          \\
diffu-GRPO           & 53.02          & 37.99          & 54.66          & 47.79          \\
wd1++ (matched)      & 50.71          & 38.06          & 53.76          & 46.73          \\
wd1++                & 52.55          & 37.60          & 53.81          & 47.25          \\
\midrule
\rowcolor{overallbg}
\textbf{Pivot-SD (ours)} & 52.87      & \textbf{39.82} & 55.42          & \textbf{48.61} \\
\bottomrule
\end{tabular}%
}
\end{table}

\subsection{What Makes Pivot-SD Work?}
\label{subsec:ablation}

\paragraph{Ablation setup.}
Table~\ref{tab:ablation-compact} isolates two design choices: the denoising state used for training and the token locations that receive credit (via cross-entropy, CE) or blame (via unlikelihood, UL).
\textbf{Positive All-token} applies CE to all supervised tokens in IG-selected states from successful trajectories. 
\textbf{Positive Pivot} applies CE only to the IG pivot from successful trajectories. 
\textbf{Entropy pivot} replaces the IG criterion with local uncertainty, selecting the top-$K$ highest-entropy committed tokens within successful trajectories and applying CE to them.
\textbf{Pos+Neg All-token} applies full-state CE/UL on IG-selected states. 
\textbf{Random-Step Pivot} applies the same pivot-local CE/UL objective to a token decoded at a random denoising step.
\textbf{Random-Token Pivot} keeps the IG-selected state but replaces the IG pivot with a randomly selected decoded token. \textbf{\ours{} (ours)} applies pivot-local CE/UL to the selected high-information pivot.

\paragraph{Pivot-local credit assignment matters.}
With both positive and negative updates, restricting the loss to the pivot token (\ours{}) beats applying it to every masked token of the selected state (Pos+Neg All-token) on all four benchmarks, by 2.5 to 5.1 points. With positive updates alone, neither target is consistently better (Positive Pivot vs.\ Positive All-token). The gain from localization therefore appears with the negative branch, consistent with state-wide unlikelihood also pushing down the correct tokens of a failed state.

\paragraph{Pivot selection matters once unlikelihood is applied.}
The two random controls keep the pivot-local CE/UL objective and change only which commitments are selected. \ours{} beats both on all four benchmarks, and Random-Step Pivot falls below the untrained base model on MATH (29.93 vs.\ 31.40). Entropy pivot changes the selection criterion in the positive-only setting, and it stays within 2.3 points of Positive Pivot on every benchmark. With positive updates alone, the choice of criterion makes little difference. With unlikelihood, applying it to random steps or random tokens costs 7.5 and 2.6 points on MATH relative to \ours{}.

\subsection{Compute Efficiency}
\label{subsec:compute}

\paragraph{Offline pivot distillation avoids online rollouts.}
Table~\ref{tab:compute} reports wall-clock training time on two GPUs. The RL baselines generate rollouts during training, which takes 3.6 to 5.4 hours at 1,000 steps and 17.3 to 22.9 hours at 5,000 steps. \ours{} samples its 800 trajectories once in 2.5 hours and fine-tunes on the cached pivots in 0.3 hours, for 2.8 hours in total. In FLOPs, which do not depend on hardware, \ours{} uses about $1.65\times$ less total compute than the budget-matched RL runs (Appendix~\ref{app:flops}).

\paragraph{Compute--performance trade-off.}
Figure~\ref{fig:acc_compute} plots average accuracy against total training time. \ours{} has the highest average accuracy while using about $8\times$ less time than the 5,000-step diffu-GRPO run and 1.3 to $1.9\times$ less than the budget-matched RL runs.

\subsection{Backbone Generalization}
\label{subsec:dream}

\paragraph{Pivot-SD transfers to another diffusion backbone.}
Table~\ref{tab:dream-compact} applies the same recipe to \dream{} \citep{ye2025dream7bdiffusionlarge}. \ours{} improves over the Dream base model and both SFT baselines on all four benchmarks, so the gains carry over to a second masked diffusion model.

\subsection{Out-of-Domain Transfer}
\label{subsec:ood-transfer}
Table~\ref{tab:ood-avg} reports accuracy on benchmarks outside the training domain. Models trained on MATH or GSM8K are evaluated on the other three benchmarks, and AceCode-trained models on MATH and GSM8K. The Overall score is the micro-average over all source-target pairs. 
\paragraph{\ours{} transfers most consistently.} 
With the same 200-question budget, \ours{} has the best overall average (48.61). It ranks first among the GSM8K-trained models, second among the AceCode-trained models, and third among the MATH-trained models, which gives it the best average rank across training domains. Appendix~\ref{sec:app-ood} lists every source-target pair.
\section{Conclusion}

We introduced Pivot-SD, a data-efficient offline self-distillation method for masked diffusion LMs that supervises only the commitments at which the model's uncertainty over the remaining positions drops most. Information gain locates these commitments, and the verifier sets the direction of the update: pivots from successful trajectories are reinforced with cross-entropy, and pivots from failed ones are penalized with unlikelihood, which leaves the valid steps around them intact. With only 200 training prompts, Pivot-SD outperforms full-sequence SFT and online RL baselines while using a fraction of their compute.

\section{Limitations}
While Pivot-SD demonstrates strong data efficiency, several aspects of the design remain open. Our objective includes a few hyperparameters that are set per configuration rather than learned, and some adjustment across domains and backbones is expected. An adaptive rule based on step-wise entropy collapse would let supervision follow task complexity without this step.
Our formulation also operates within a denoising block, so extending step-conditioned credit assignment across block boundaries would be needed for long-context reasoning. Finally, our experiments span two masked diffusion backbones at a comparable scale, and applying pivot-local supervision as dLMs grow is a natural next step.

\section*{Acknowledgements}
RGK gratefully acknowledges support from the Canada Research Chairs Program (CRC-2022-00049) and the Canada CIFAR AI Chairs Program. This research was funded in part by an NFRF Special Call Award (NFRFR-2022-00526) and NSERC Discovery Grant (RGPIN-2022-04546). This work was also partly supported by the ``Advanced GPU Utilization Support Program'' funded by the Government of the Republic of Korea (Ministry of Science and ICT) (05-26-04-0100), and by Institute for Information \& Communications Technology Planning \& Evaluation (IITP) grants funded by the Korea government (MSIT) (RS-2022-00143911, AI Excellence Global Innovative Leader Education Program; RS-2026-25522672, Development of Unified Reasoning Technology Mimicking Human Cognition for Hierarchical Understanding and Unbounded Problem Solving; RS-2019-II190075, Artificial Intelligence Graduate School Support Program (KAIST)).

\bibliography{custom}

\clearpage

\appendix

\section{Credit-Assignment Landscape}
\label{app:credit-landscape}

Table~\ref{tab:credit-landscape-full} compares Pivot-SD with related methods, including process reward models, the training-free Info-Gain Sampler, and answer-aware selection \citep{horvitz2026rethinkingreasoningmdlmsearly, yang2026improvingsamplingmaskeddiffusion}. Two properties distinguish Pivot-SD in this comparison. It trains only on selected realized commitments $(t, p, y_p)$ under the partially denoised state $M_t$, and it builds localized negative targets from failed trajectories. None of the other methods in the table combines the two.

\begin{table*}[t]
\centering
\small
\setlength{\tabcolsep}{5pt}
\renewcommand{\arraystretch}{1.3}
\caption{\textbf{Credit-assignment landscape.} Among the methods listed, Pivot-SD is the only one that trains on selected realized commitments $(t,p,y_p)$ alone and builds a localized negative target from failed trajectories.}
\begin{tabular}{@{}p{0.17\linewidth}p{0.19\linewidth}p{0.22\linewidth}p{0.13\linewidth}p{0.16\linewidth}@{}}
\toprule
\textbf{Method} & \textbf{Local unit} & \textbf{Localization signal} & \textbf{Training regime} & \textbf{Localized Failure Target} \\
\midrule
STaR / rejection SD \citep{NEURIPS2022_639a9a17} &
Full successful rationale &
Final correctness &
Offline SFT &
No (failed traces are filtered or retried) \\

Process supervision (PRMs) \citep{lightman2024lets, wang-etal-2024-math} &
Textual reasoning step &
Human or separately constructed process label &
PRM / SFT / RL &
Possible, but requires process-label construction \\

GIFT \citep{xu2026giftguidedimportanceawarefinetuning} &
Final-response token position &
Local predictive entropy &
Offline weighted SFT &
No \\

Denoising Process Reward / ATPO \citep{xie-etal-2026-advancing, chen2025reasoningdiffusionlargelanguage} &
Denoising interval or step &
Outcome-progress reward or trajectory uncertainty &
Online RL &
No explicit token-level UL target \\

AGRPO \citep{zhan2026simplepolicygradientsreasoning} &
Sampled unmasking step &
Trajectory-level advantage; steps sampled uniformly or by entropy &
Online RL &
No (trajectory-level advantage on every sampled step) \\

DiSPO \citep{oba2026diffusionstatepolicyoptimizationmasked} &
State-conditioned mask filling / newly filled tokens &
Branched terminal rewards &
Policy optimization &
No cached realized-pivot UL target \\

Info-Gain Sampler / KLASS \citep{yang2026improvingsamplingmaskeddiffusion, NEURIPS2025_8514c1f1} &
Candidate inference action &
Effect on future-mask uncertainty &
Inference only &
N/A \\

Horvitz et al. \citep{horvitz2026rethinkingreasoningmdlmsearly} &
Answer block / reasoning trace &
Answer entropy or gold-answer probability &
Inference / trace self-distillation &
No localized failed-pivot UL \\

\textbf{Pivot-SD (ours)} &
Selected realized commitment $(t,p,y_p)$ under $M_t$ &
Entropy drop over the still-masked positions across the realized step &
Cached offline CE / UL &
\textbf{Yes} \\
\bottomrule
\end{tabular}

\label{tab:credit-landscape-full}
\end{table*}

\section{Illustration of Information-Gain Pivot Selection}
\label{sec:appendix_trajectory_analysis}
Figure~\ref{fig:pivot_sd_main} illustrates pivot selection. For each
denoising step, \ours{} measures how much the entropy of the positions that remain
masked drops once the step's commitments are written. The commitments at the
highest-scoring steps become pivots $(t, p, y_p)$, and the CE or UL update is
applied only at those positions.

\section{Experimental Details and Configurations}
\label{sec:app-experimental-details}

This section lists the training and evaluation configurations.

\paragraph{Hardware Infrastructure.}
All training, trajectory generation, and evaluation ran on two NVIDIA L40 GPUs.

\paragraph{Generation and Rollout Limits.}
All training-time generation uses a maximum length of \textbf{256 tokens}. This covers the offline trajectory sampling for \ours{}, the success-only sampling for SFT-SD, and the online rollouts of diffu-GRPO and wd1++, so every method trains under the same response budget.
Generation uses the model's default confidence-based sampler with a block length of 64 and one denoising step per token, so exactly one position is committed at each step. Information gain is therefore scored once per step and inherited by the single token committed there.
Two classes of commitments are excluded from the pivot pool before ranking: steps at a block boundary, where $H_{\mathrm{post}}$ would span a different set of positions, and end-of-sequence or turn-end tokens, which carry no reasoning content.

\paragraph{Optimization and Hyperparameters.}
All offline methods (SFT-GT, SFT-SD, and \ours{}) are trained for 1,000 optimizer steps with a batch size of 8. The baseline RL methods are trained under both a budget-matched setting (1,000 steps) and an extended setting (5,000 steps) to observe convergence scaling. For \ours{}, rather than tuning the negative unlikelihood weight $\lambda_{\mathrm{neg}}$, we set it from the positive-to-negative trajectory ratio of the offline pool so that the two branches contribute comparably. We set $\lambda_{\mathrm{neg}} = 1$ for MATH, and $\lambda_{\mathrm{neg}} = 5$ for GSM8K to account for an approximate 5:1 ratio of positive to negative trajectories. Coding is the one exception: since models are trained on AceCode but evaluated on HumanEval+ and MBPP+, the ratio-matched value risks over-suppression under distribution shift, so we lower the penalty to $\lambda_{\mathrm{neg}} = 0.02$ along a log scale on a held-out split of roughly 100 examples.

\paragraph{Evaluation Protocol and Decoding Strategy.}
We evaluate on MATH500, GSM8K, HumanEval+, and MBPP+ with maximum generation lengths of \textbf{256 tokens}, which matches training, and \textbf{512 tokens}, which tests length transfer. We evaluate checkpoints at 500 and 1,000 optimizer steps. All \llada{} evaluations use top-1 confidence decoding at temperature $0$, so evaluation is deterministic.

\section{Multi-Seed Evaluation Protocol}
\label{app:multiseed}

\paragraph{Scope.}
We repeat every method trained under our controlled budget (200 questions, 1{,}000 optimizer steps) three times and report means in the main tables. The base model is not trained and is evaluated with deterministic decoding (temperature $0$, top-1), so seed variance is undefined for it. The extended-budget RL baselines are too slow to repeat three times and are reported as single runs. The 512-token columns are likewise single runs, since they measure length transfer rather than the training-matched setting.

\paragraph{What is re-randomized.}
Each run independently re-samples (i) the 200-question training set, (ii) the trajectory pool, re-generated from the frozen base model at temperature $0.9$, and (iii) the training seed. The reported variance therefore covers the whole offline pipeline that produces $\mathcal{D}_{\text{pivot}}$, including question selection and trajectory sampling. Evaluation is deterministic, so all of it comes from training.

\paragraph{Reading the standard deviations.}
The code benchmarks are small, so their scores are coarse. HumanEval+ contains 164 problems, and one problem corresponds to 0.61 points. A standard deviation of 2.75 points on HumanEval+ therefore reflects a difference of four to five problems between runs. With three runs per method, the standard deviations are themselves rough estimates.

\paragraph{Separation between methods.}
We report means and standard deviations and do not claim statistical significance. On MATH, GSM8K, and HumanEval+, the gap between Pivot-SD and the strongest competing method (2.0, 2.6, and 4.7 points) exceeds the standard deviation of either method. On MBPP+, the 0.8-point advantage over SFT-GT is small relative to run-to-run variability ($47.12 \pm 1.20$ vs.\ $46.32 \pm 0.45$). Table~\ref{tab:multiseed-ablation} also shows that the random-selection controls vary widely across runs. Random-Step Pivot has a standard deviation of $3.13$ on MATH, where its mean is below the untrained base model, and $5.51$ on HumanEval+.

\begin{table*}[t]
\centering
\small
\caption{\textbf{Main comparison with per-run standard deviations.} Mean $\pm$
standard deviation over three fully re-randomized runs, 256-token evaluation.
Rows correspond to the $\ddag$-marked entries of Table~\ref{tab:llada-main}.
RL baselines are the budget-matched variants.}
\label{tab:multiseed-main}
\begin{tabular}{lcccc}
\toprule
\textbf{Method} & \textbf{MATH} & \textbf{GSM8K} & \textbf{HumanEval+} & \textbf{MBPP+} \\
\midrule
SFT-GT      & 35.47 \std{0.23} & 76.94 \std{1.05} & 33.94 \std{2.32} & 46.32 \std{0.45} \\
SFT-SD      & 31.73 \std{0.58} & 73.90 \std{0.89} & 35.78 \std{1.54} & 45.92 \std{1.46} \\
diffu-GRPO  & 33.47 \std{1.03} & 76.04 \std{0.80} & 32.52 \std{1.27} & 43.39 \std{0.27} \\
wd1++       & 33.27 \std{0.50} & 76.93 \std{0.59} & 33.54 \std{2.80} & 43.12 \std{0.27} \\
\midrule
\rowcolor{overallbg}
\textbf{\ours{} (ours)} & \textbf{37.47} \std{1.21} & \textbf{79.51} \std{1.30}
            & \textbf{40.43} \std{2.75} & \textbf{47.12} \std{1.20} \\
\bottomrule
\end{tabular}
\end{table*}

\begin{table*}[t]
\centering
\small
\caption{\textbf{Ablation with per-run standard deviations.} Mean $\pm$ standard
deviation over three fully re-randomized runs, 256-token evaluation.}
\label{tab:multiseed-ablation}
\begin{tabular}{lcccc}
\toprule
\textbf{Method} & \textbf{MATH} & \textbf{GSM8K} & \textbf{HumanEval+} & \textbf{MBPP+} \\
\midrule
Pos+Neg All-token  & 35.00 \std{1.25} & 74.44 \std{0.70} & 36.80 \std{0.35} & 44.54 \std{2.37} \\
Random-Step Pivot  & 29.93 \std{3.13} & 76.51 \std{0.74} & 39.00 \std{5.51} & 44.80 \std{0.85} \\
Random-Token Pivot & 34.87 \std{1.94} & 76.47 \std{1.68} & 34.78 \std{1.07} & 45.24 \std{0.70} \\
Entropy pivot      & 32.27 \std{1.10} & 74.75 \std{0.82} & 35.57 \std{1.76} & 44.80 \std{1.10} \\
\midrule
\rowcolor{overallbg}
\textbf{\ours{} (ours)} & \textbf{37.47} \std{1.21} & \textbf{79.51} \std{1.30}
            & \textbf{40.43} \std{2.75} & \textbf{47.12} \std{1.20} \\
\bottomrule
\end{tabular}
\end{table*}

\paragraph{Separation between methods.}
On MATH, GSM8K, and HumanEval+, the one-standard-deviation intervals of \ours{} and of the strongest competing method do not overlap. On MBPP+ they overlap ($47.12 \pm 1.20$ vs.\ $46.32 \pm 0.45$ for SFT-GT), so we count MBPP+ as a tie. Table~\ref{tab:multiseed-ablation} also shows that the random-selection controls vary widely across runs. Random-Step Pivot has a standard deviation of $3.13$ on MATH, where its mean is below the untrained base model, and $5.51$ on HumanEval+.

\section{Training Budget Sweep}
\label{sec:app-budget}

Our main experiments fix the training budget at $q=200$ questions. To verify that the comparison is not specific to that choice, we vary $q \in \{50, 200, 400, 2000\}$ and retrain every method under the same protocol in Table~\ref{tab:budget-sweep}.

\begin{table*}[t]
\centering
\small
\setlength{\tabcolsep}{4pt}
\caption{\textbf{Accuracy across training budgets} (256-token evaluation).
Column headers give the number of training questions $q$. Step budgets are fixed within each group ($q=50$/$200$: 1{,}000 steps; $q=400$: 2{,}000; $q=2000$: 5{,}000), matching the extended RL baselines. All entries are single runs. Multi-seed means for $q=200$ are in Table~\ref{tab:multiseed-main}.}
\label{tab:budget-sweep}
\begin{tabular}{l cccc cccc cccc cccc}
\toprule
& \multicolumn{4}{c}{\textbf{MATH}}
& \multicolumn{4}{c}{\textbf{GSM8K}}
& \multicolumn{4}{c}{\textbf{HumanEval+}}
& \multicolumn{4}{c}{\textbf{MBPP+}} \\
\cmidrule(lr){2-5} \cmidrule(lr){6-9} \cmidrule(lr){10-13} \cmidrule(lr){14-17}
\textbf{Method}
& 50 & 200 & 400 & 2000
& 50 & 200 & 400 & 2000
& 50 & 200 & 400 & 2000
& 50 & 200 & 400 & 2000 \\
\midrule
SFT-GT      & 36.0 & 35.6 & 34.2 & 35.0 & 76.7 & 75.8 & 77.9 & 75.1 & 32.9 & 32.9 & 32.9 & 33.5 & \textbf{46.6} & \textbf{46.6} & 45.8 & 46.8 \\
SFT-SD      & 31.6 & 31.4 & 33.0 & 31.6 & 72.4 & 74.9 & 72.7 & 73.2 & 37.8 & 35.4 & 37.2 & 37.8 & 45.5 & 45.8 & 44.7 & 45.5 \\
diffu-GRPO  & 34.0 & 34.6 & 34.6 & 35.2 & 76.4 & 75.7 & 76.4 & 77.6 & 31.7 & 31.1 & 32.9 & 34.2 & 42.3 & 43.7 & 43.9 & 42.6 \\
wd1++       & 32.4 & 33.8 & \textbf{39.2} & 35.4 & 75.9 & 77.1 & 79.0 & 76.7 & 32.3 & 32.9 & 31.1 & 34.2 & 44.7 & 43.1 & 45.8 & 45.0 \\
\midrule
\rowcolor{overallbg}
\textbf{\ours{} (ours)}
            & \textbf{36.2} & \textbf{38.6} & 36.0 & \textbf{37.2} & 75.0 & \textbf{79.5} & \textbf{80.1} & \textbf{78.9} & \textbf{42.1} & \textbf{40.9} & \textbf{40.2} & \textbf{38.4} & 46.0 & 46.0 & \textbf{46.6} & \textbf{47.1} \\
\bottomrule
\end{tabular}
\end{table*}

\paragraph{The advantage holds at the largest budget.}
diffu-GRPO and wd1++ keep generating rollouts during training, so the largest budget gives online exploration the most room to help. At $q=2000$, where they train for 5,000 steps, \ours{} is still best on all four benchmarks.

\paragraph{Small budgets already beat large ones.}
With 200 questions, \ours{} matches or exceeds every baseline at every budget on GSM8K and HumanEval+, including baselines trained on ten times the data.

\paragraph{Exceptions.}
Baselines lead in four cases. In three of them the margin is at most 1.7 points: GSM8K at $q=50$ and MBPP+ at $q=50$ and $q=200$. The one larger gap is a wd1++ spike on MATH at $q=400$ (39.2). wd1++ does not reach this value at $q=2000$ (35.4) or in the multi-seed runs at $q=200$ (33.27, Table~\ref{tab:multiseed-main}).

\section{Pivot Budget $K$}
\label{sec:app-ksweep}

The pivot budget $K$ is the only quantity in \ours{} that controls how much of a trajectory receives supervision. We fixed $K=10$ before running experiments. Table~\ref{tab:k-sweep} reports the sweep $K \in \{5, 10, 20, 40, 80\}$ for \ours{} and for the two random-selection controls, all trained on the same trajectory pool.

\begin{table*}[t]
\centering
\small
\setlength{\tabcolsep}{5pt}
\caption{\textbf{Effect of the pivot budget $K$} ($q=200$, 256-token evaluation). $K$ is the number of pivots supervised per trajectory. Random-Step and Random-Token are the two random-selection controls of Table~\ref{tab:ablation-compact}. We use $K=10$. $K=40$ and $80$ show what happens when selection is no longer selective. Entries are single runs on a shared trajectory pool. The multi-seed mean for $K=10$ is in Table~\ref{tab:multiseed-main}.}
\label{tab:k-sweep}
\begin{tabular}{c ccc ccc ccc ccc}
\toprule
& \multicolumn{3}{c}{\textbf{MATH}}
& \multicolumn{3}{c}{\textbf{GSM8K}}
& \multicolumn{3}{c}{\textbf{HumanEval+}}
& \multicolumn{3}{c}{\textbf{MBPP+}} \\
\cmidrule(lr){2-4} \cmidrule(lr){5-7} \cmidrule(lr){8-10} \cmidrule(lr){11-13}
$K$
& \shortstack{Random\\Step} & \shortstack{Random\\Token} & \textbf{Ours}
& \shortstack{Random\\Step} & \shortstack{Random\\Token} & \textbf{Ours}
& \shortstack{Random\\Step} & \shortstack{Random\\Token} & \textbf{Ours}
& \shortstack{Random\\Step} & \shortstack{Random\\Token} & \textbf{Ours} \\
\midrule
5  & 22.8 & 34.2 & \textbf{37.8} & 74.3 & 75.1 & \textbf{79.6} & \textbf{39.0} & 34.8 & \textbf{39.0} & 42.6 & 44.2 & \textbf{45.0} \\
10 & 26.6 & 33.2 & \textbf{38.6} & 76.9 & 75.7 & \textbf{79.5} & 35.4 & 37.8 & \textbf{40.9} & \textbf{46.0} & 45.8 & \textbf{46.0} \\
20 & 17.6 & 36.4 & \textbf{38.2} & 73.8 & 76.0 & \textbf{76.8} & 39.6 & 37.2 & \textbf{41.5} & 41.5 & \textbf{45.5} & \textbf{45.5} \\
40 & 22.0 & \textbf{34.6} & 22.8 & 73.0 & 77.0 & \textbf{79.5} & 39.0 & \textbf{43.3} & 39.6 & 41.0 & \textbf{44.4} & 39.7 \\
80 & 20.4 & \textbf{33.0} & 17.6 & 69.8 & \textbf{77.6} & 76.7 & 37.2 & \textbf{42.7} & 30.5 & 34.7 & \textbf{41.8} & 37.6 \\
\bottomrule
\end{tabular}
\end{table*}

\paragraph{$K=10$ sits inside a flat plateau.}
\ours{} is best or tied for best at every $K \in \{5, 10, 20\}$ on all four benchmarks.

\paragraph{Large $K$ removes selectivity.}
At $K=40$, MATH falls to 22.8 and both code benchmarks drop, while GSM8K is unchanged. A large $K$ admits steps that barely change the remaining entropy, so unlikelihood reaches positions that did not shape the trajectory.

\paragraph{Sparsity alone does not explain the gain.}
Random-Step Pivot supervises exactly as many positions as \ours{} but collapses on MATH for $5 \le K \le 20$ (17.6 to 26.6, against 37.8 to 38.6 for \ours{}).

\section{Compute and FLOPs Accounting}
\label{app:flops}

Table~\ref{tab:compute} and Figure~\ref{fig:acc_compute} report wall-clock time, which depends on hardware and implementation. Table~\ref{tab:flops} adds forward-and-backward FLOPs, model calls, and the number of supervised tokens.

\paragraph{Setup.}
All quantities use a common sequence length $L = 512$ (maximum prompt length plus a $256$-token generation budget). Generation FLOPs are estimated as $\mathrm{NFE} \times 2NL$ for a model with $N = 8\times10^{9}$ parameters, where NFE is the number of function evaluations (forward passes). \ours{} samples $800 = 200\times 4$ offline trajectories with $256$ sampler steps each, for $800\times 256 = 204{,}800$ NFE. The information-gain scores reuse these forward passes (\S\ref{sec:information-gain-pivot-selection}), so pivot selection adds no NFE. The budget-matched RL runs train for $1{,}000$ optimizer steps and refresh their completions every $12$ steps ($\texttt{num\_iterations}{=}12$), which gives $84 = \lfloor(1000{-}1)/12\rfloor + 1$ generation rounds at steps $0,12,\ldots,996$. Each round produces $16 = 2\times 2\times 4$ rollouts ($2$ GPUs, $2$ prompts per device, $4$ generations per prompt), for $84\times 16 = 1{,}344$ rollouts and $1{,}344\times 256 = 344{,}064$ rollout NFE. Adding the policy and scoring passes gives $358{,}080$ online NFE. The $200$-question set is the pool of unique prompts that these rounds resample from. RL numbers are averaged over the three training domains. The supervised-token count includes only positions that receive a loss.

\begin{table*}[t]
\centering
\small
\renewcommand{\arraystretch}{1.2}
\caption{\textbf{Compute accounting} at $L=512$ and $N=8\times 10^9$. \ours{} uses about $1.65\times$ fewer total FLOPs than the budget-matched RL runs. Its generation happens once before training, while the RL runs generate rollouts throughout training.}
\label{tab:flops}
\begin{tabular}{lccc}
\toprule
\textbf{Metric} & \textbf{Pivot-SD (ours)} & \textbf{diffu-GRPO} & \textbf{wd1++} \\
\midrule
Generation (PFLOPs) & \textbf{1{,}678} (offline) & 2{,}819 ($1.68\times$) & 2{,}819 \\
Training fwd+bwd (PFLOPs) & \textbf{197} & 274 & 274 \\
\textbf{Total Compute (PFLOPs)} & \textbf{1{,}875} & 3{,}093 ($\sim$$1.65\times$) & 3{,}093 \\
\midrule
Model Calls (NFE) & \textbf{204{,}800} (offline) & 358{,}080 (online) & 358{,}080 (online) \\
Supervised Tokens / Trajectory & \textbf{10} (3.9\%) & 256 (100\%) & 256 (100\%) \\
\midrule
Wall-Clock Time (2$\times$L40) & \textbf{2.8 h} & 5.4 h & 3.6 h \\
\bottomrule
\end{tabular}
\end{table*}

\paragraph{FLOPs.}
Total compute is $1{,}875$ PFLOPs for \ours{} and $3{,}093$ PFLOPs for each budget-matched RL run, a ratio of about $1.65\times$. Most of the difference comes from generation ($1{,}678$ vs.\ $2{,}819$ PFLOPs). \ours{} supervises $10$ tokens per trajectory, $3.9\%$ of the response. Its trajectories do not depend on each other, so generation parallelizes across prompts and devices, whereas the RL generation rounds must run in sequence.

\paragraph{Hyperparameter search.}
The learning rate and adapter settings are taken from the RL baselines. $K=10$ was fixed before experiments (Appendix~\ref{sec:app-ksweep}), and $\lambda_{\text{neg}}$ is set from the positive-to-negative trajectory ratio for MATH and GSM8K without search. The only search is for code. Models are trained on AceCode and evaluated on HumanEval+ and MBPP+, and under this distribution shift the ratio-matched value can over-suppress. We therefore screened $\lambda_{\text{neg}}$ on a logarithmic scale using a held-out split of about $100$ examples, a one-time search of about $2$ hours that can run in parallel.

\section{Out-of-Domain Generalization}
\label{sec:app-ood}

Table~\ref{tab:ood-generalization} evaluates models trained on one domain on the benchmarks of the other domains.

\begin{table*}[h!]
\centering
\small
\renewcommand{\arraystretch}{1.08}
\caption{
\textbf{Out-of-Domain (OOD) generalization results.} Models are trained on one domain and evaluated on others. ``budget-matched'' denotes models trained with the 1,000-step budget configuration matching \ours{}.
}
\label{tab:ood-generalization}
\resizebox{\textwidth}{!}{%
\begin{tabular}{l cccc cccc ccc}
\toprule
\multirow{2}{*}{\textbf{Method}} 
& \multicolumn{4}{c}{\textbf{MATH Trained}} 
& \multicolumn{4}{c}{\textbf{GSM8K Trained}} 
& \multicolumn{3}{c}{\textbf{AceCode Trained}} \\
\cmidrule(lr){2-5} \cmidrule(lr){6-9} \cmidrule(lr){10-12}
& GSM8K & HumanEval+ & MBPP+ & Avg. & MATH & HumanEval+ & MBPP+ & Avg. & MATH & GSM8K & Avg. \\
\midrule

SFT-GT 
& 76.35 & 37.80 & 46.56 & 53.57 
& 35.40 & 37.20 & 45.77 & 39.46 
& 30.20 & 73.39 & 51.80 \\

SFT-SD 
& 75.74 & 32.93 & 44.71 & 51.13 
& 30.60 & 30.49 & 45.24 & 35.44 
& 29.40 & 73.01 & 51.21 \\

diffu-GRPO (budget-matched) 
& 75.21 & 33.54 & 43.92 & 50.89 
& 33.80 & 31.10 & 44.44 & 36.45 
& 34.40 & 76.57 & 55.49 \\

diffu-GRPO 
& 76.35 & 37.20 & 45.50 & 53.02 
& 35.80 & 32.93 & 45.24 & 37.99 
& 33.80 & 75.51 & 54.66 \\

wd1++ (budget-matched) 
& 76.42 & 32.32 & 43.39 & 50.71 
& 35.00 & 36.59 & 42.59 & 38.06 
& 32.60 & 74.91 & 53.76 \\

wd1++ 
& 78.09 & 35.37 & 44.18 & 52.55 
& 35.60 & 33.54 & 43.65 & 37.60 
& 32.40 & 75.21 & 53.81 \\

\midrule
\rowcolor{overallbg}
\textbf{\ours{} (ours)}
& 78.09 & \textbf{36.59} & 43.92 & \textbf{52.87} 
& \textbf{36.60} & \textbf{38.41} & 44.44 & \textbf{39.82} 
& \textbf{34.80} & \textbf{76.04} & \textbf{55.42} \\

\midrule
\multicolumn{12}{l}{\textit{Ablation Variants (Budget-matched)}} \\
\midrule
Positive Pivot
& 73.46 & 33.54 & 41.27 & 49.42 
& 31.40 & 36.59 & 41.01 & 36.33 
& 30.80 & 73.92 & 52.36 \\

Positive All-token
& 75.36 & 34.76 & 42.86 & 50.99 
& 29.60 & 36.59 & 43.92 & 36.70 
& 32.60 & 73.77 & 53.19 \\

Pos+Neg All-token
& 75.44 & 38.41 & 43.39 & 52.41 
& 33.00 & 34.76 & 44.44 & 37.40 
& 33.60 & 75.44 & 54.52 \\

Random-Token Pivot
& 77.10 & 36.59 & 46.03 & 53.24 
& 31.80 & 37.80 & 45.77 & 38.46 
& 34.60 & 74.07 & 54.34 \\

Random-Step Pivot
& 75.06 & 32.93 & 37.57 & 48.52 
& 33.40 & 29.27 & 41.27 & 34.65 
& 35.00 & 76.42 & 55.71 \\

\bottomrule
\end{tabular}%
}
\end{table*}

\ours{} has the best average among the GSM8K-trained models and the best overall micro-average (Table~\ref{tab:ood-avg}). For the MATH- and AceCode-trained models, its average is within 0.7 points of the best.

\section{Reward Sparsity in wd1++ and the Bootstrap Reward}
\label{sec:appendix_bootstrap}

On the 200-prompt AceCode subset ($s200$), the \texttt{wd1++} baseline stopped learning. 

The base model often failed every unit test for all samples in a batch, so the group reward, and with it the advantage, was zero. \texttt{wd1++} combines a positively weighted likelihood term with a negatively weighted unlikelihood term. With zero advantage the weights become uniform and the two gradients cancel. The checkpoints at steps 500 and 1000 are identical (``w/o bootstrap'' in Table~\ref{tab:bootstrap_results}).

To give \texttt{wd1++} a learning signal in this setting, we added a \textbf{bootstrap reward} to the $s200$ runs. When the unit-test pass rate was zero, a completion received a small structural reward (up to $+0.05$) if it met basic syntactic requirements, such as a proper \texttt{def} header, a valid \texttt{return} statement, and overall syntactic validity.

\begin{table}[ht]
\centering
\small
\setlength{\tabcolsep}{3pt}
\caption{\textbf{Effect of the bootstrap reward on \texttt{wd1++} trained on 200 AceCode prompts.} GSM and HE+ denote GSM8K and HumanEval+. Without bootstrap, training stalls between steps 500 and 1000 because the zero-advantage gradients cancel. With the bootstrap reward, training resumes on code and MBPP+ gains 3.97 points by step 1000, while MATH does not improve. The \texttt{wd1++} numbers in the main tables (Tables~\ref{tab:llada-main} and~\ref{tab:multiseed-main}) come from the standard runs without bootstrap shaping. The $47.09$ here is the single-run peak after bootstrap shaping and is outside the multi-seed protocol.}
\resizebox{\columnwidth}{!}{%
\begin{tabular}{l cc cc cc}
\toprule
 & \multicolumn{2}{c}{\textbf{Math}} & \multicolumn{2}{c}{\textbf{Code (256)}} & \multicolumn{2}{c}{\textbf{Code (512)}} \\
\cmidrule(lr){2-3} \cmidrule(lr){4-5} \cmidrule(lr){6-7}
\textbf{Variant} & \textbf{MATH} & \textbf{GSM} & \textbf{MBPP+} & \textbf{HE+} & \textbf{MBPP+} & \textbf{HE+} \\
\midrule
\multicolumn{7}{@{}l}{\textit{w/o bootstrap}} \\
\quad Step 500  & 32.60 & 74.91 & 43.12 & 32.93 & 43.92 & 40.24 \\
\quad Step 1000 & -     & -     & 43.12 & 32.93 & -     & -     \\
\midrule
\multicolumn{7}{@{}l}{\textit{w/ bootstrap}} \\
\quad Step 500  & 33.20 & 74.53 & 44.71 & 31.10 & 46.56 & 39.63 \\
\quad Step 1000 & 31.20 & 74.45 & 47.09 & 33.54 & 46.03 & 40.85 \\
\bottomrule
\end{tabular}%
}
\label{tab:bootstrap_results}
\end{table}

With the bootstrap reward, \texttt{wd1++} resumes learning on code and gains 3.97 points on MBPP+ at 256 tokens by step 1000 (Table~\ref{tab:bootstrap_results}). At 512 tokens it gains 2.1 points on MBPP+ and 0.6 on HumanEval+.. 

Even with this reward, \texttt{wd1++} does not surpass \ours{}. On MBPP+ the shaped run reaches $47.09$, close to Pivot-SD ($47.12 \pm 1.20$). On HumanEval+, \ours{} leads by 6.9 points ($40.43$ vs.\ $33.54$). This \texttt{wd1++} run is trained on AceCode, so its MATH and GSM8K scores are out-of-domain. Against \ours{} trained on the same AceCode subset, \ours{} is higher on both ($34.80$ vs.\ $31.20$ on MATH and $76.04$ vs.\ $74.45$ on GSM8K, Table~\ref{tab:ood-generalization}). \ours{} reaches these results without any reward shaping.

\section{Artifact Licenses and Intended Use}
\label{sec:appendix_licenses}

In our experiments, we utilize publicly available mathematical and code generation benchmarks, including GSM8K (MIT License), MATH (MIT License), HumanEval (MIT License), and MBPP (CC-BY-4.0). We also use two pre-trained masked diffusion language models: LLaDA-8B-Instruct \citep{nie2026large}, released under the MIT License, and Dream-v0-Instruct-7B \citep{ye2025dream7bdiffusionlarge}, released under the Apache 2.0 License.

Our use of these artifacts is consistent with their intended use as standard structural reasoning and evaluation benchmarks for language modeling research. These datasets do not contain personally identifying information or offensive content. Furthermore, the trajectories and evaluation scripts generated during the development of Pivot-SD will be open-sourced under the MIT License to facilitate reproducibility and future research in masked diffusion language models.

\section{Potential Risks}
\label{sec:potential-risks}

\ours{} carries the following risks:

\begin{itemize}[leftmargin=*]
    \item \textbf{Bias and Hallucination Amplification:} As an offline self-distillation framework, \ours{} relies on trajectories generated by the frozen base model. Filtering by outcome does not remove biases or hallucinations present in the base model, and training on its own trajectories can amplify them.
    \item \textbf{Dependence on Automated Verifiers:} Our trajectory labeling currently relies on exact-match checking for mathematics and unit-test execution for code. Applying this framework to open-ended generation tasks without clear objective verifiers may lead to reward hacking or optimization toward superficial proxies of quality.
\end{itemize}

\section{Broader Impacts}

\ours{} lowers the compute needed to improve reasoning in masked diffusion language models. In our setting it trained in about one-eighth of the wall-clock time of the 5,000-step diffu-GRPO run, using 200 prompts and a single offline sampling pass. This makes post-training research on diffusion language models feasible on modest academic hardware.

Localized unlikelihood may also be useful for alignment. Rejection fine-tuning and full-sequence unlikelihood suppress valid fragments together with the undesired behavior, whereas \ours{} penalizes only selected commitments. Stronger reasoning and code generation carry dual-use risks, such as generating malicious code or persuasive misinformation. Because \ours{} relies on an external verifier and trains offline on a fixed dataset, existing safety filters can be applied to that dataset before training.

\end{document}